\documentclass[conference]{IEEEtran}

\IEEEoverridecommandlockouts
\usepackage{cite}
\usepackage{amsmath,amssymb,amsfonts}
\usepackage{algorithmic}
\usepackage{graphicx}
\usepackage{textcomp}
\usepackage{xcolor}
\usepackage{bbm}
\usepackage{booktabs}
\usepackage{multirow}
\def\BibTeX{{\rm B\kern-.05em{\sc i\kern-.025em b}\kern-.08em
    T\kern-.1667em\lower.7ex\hbox{E}\kern-.125emX}}
\begin{document}

\title{Unifying Prediction and Explanation in Time-Series Transformers via Shapley-based Pretraining\\
}



\author{
    \IEEEauthorblockN{Qisen Cheng\IEEEauthorrefmark{1}, Jinming Xing\IEEEauthorrefmark{2}, Chang Xue\IEEEauthorrefmark{3}, Xiaoran Yang\IEEEauthorrefmark{2}}
    \IEEEauthorblockA{\IEEEauthorrefmark{1}Department of Electrical Engineering and Computer Science, University of Michigan, Ann Arbor, MI, USA}
    \IEEEauthorblockA{\IEEEauthorrefmark{2}School of Computer Science, North Carolina State University, Raleigh, NC, USA
    }
    \IEEEauthorblockA{\IEEEauthorrefmark{3}Katz School of Science and Health, Yeshiva University, New York, NY, USA
    }
    qisench@umich.edu, \{jxing6, xyang49\}@ncsu.edu, cxue@mail.yu.edu
}


\maketitle

\begin{abstract}
In this paper, we propose ShapTST, a framework that enables time-series transformers to efficiently generate Shapley-value-based explanations alongside predictions in a single forward pass. Shapley values are widely used to evaluate the contribution of different time-steps and features in a test sample, and are commonly generated through repeatedly inferring on each sample with different parts of information removed. Therefore, it requires expensive inference-time computations that occur at every request for model explanations. In contrast, our framework unifies the explanation and prediction in training through a novel Shapley-based pre-training design, which eliminates the undesirable test-time computation and replaces it with a single-time pre-training. Moreover, this specialized pre-training benefits the prediction performance by making the transformer model more effectively weigh different features and time-steps in the time-series, particularly improving the robustness against data noise that is common to raw time-series data. We experimentally validated our approach on eight public datasets, where our time-series model achieved competitive results in both classification and regression tasks, while providing Shapley-based explanations similar to those obtained with post-hoc computation. Our work offers an efficient and explainable solution for time-series analysis tasks in the safety-critical applications.

\end{abstract}


\section{Introduction}
Time-series classification and regression are fundamental data analysis tasks in various business, such as finance \cite{sezer2020financial, li2021augmented}, healthcare \cite{yao2017deepsense, liu2024analysis, yi2022attention} and manufacturing \cite{cheng2022estimation, zhang2022predicting, balakrishnan20236}. Therefore, it has received significant attention from the deep learning community \cite{ismail2019deep}. Time-series Transformer models (TST) have recently gained prominence and achieved remarkable success \cite{wen2022transformers, li2024enhancing} in the time-series domain. However, existing TST models are often considered black-box solutions, lacking interpretability, which limits their utility and reliability, particularly for safety-critical applications \cite{ismail2020benchmarking, pan202432, deng2021short}. This limitation hinders their deployment in production environments. Thus, it is desirable to integrate an explanation mechanism into the TST model and generate explanations during inference.

A popular approach to interpret time-series models is through Shapley-based explanations, which generate Shapley values to quantify the contribution of each element—features or time steps—in the input time-series data \cite{mishra2017local, lundberg2017unified, covert2021improving}. Rooted in credit assignment theory, this approach calculates Shapley values by evaluating the difference in the model’s output when a specific part of the input is included versus when it is omitted, effectively measuring the marginal contribution of that element to the overall prediction \cite{hamilton2020time}. Among the various Shapley-based explanation techniques, TimeShap \cite{bento2021timeshap} is a method specifically designed for time-series data, offering both feature-wise and time-step explanations through multidimensional perturbations and sampling. While this approach is highly effective, it necessitates repeated inference with multiple forward passes, where each pass processes a sample with certain information removed. This inherently makes the explanation generation process computationally intensive and slow \cite{smagulova2019survey}. Furthermore, since these computations must be performed for every new request for model explanations, they impose a significant computational burden after deployment. Ideally, a deployed model should produce both predictions and corresponding explanations simultaneously, without incurring additional inference-time computations or compromising prediction quality. However, achieving such integration remains an unresolved challenge in time-series tasks.

\begin{figure*}[h]
\includegraphics[width=\textwidth]{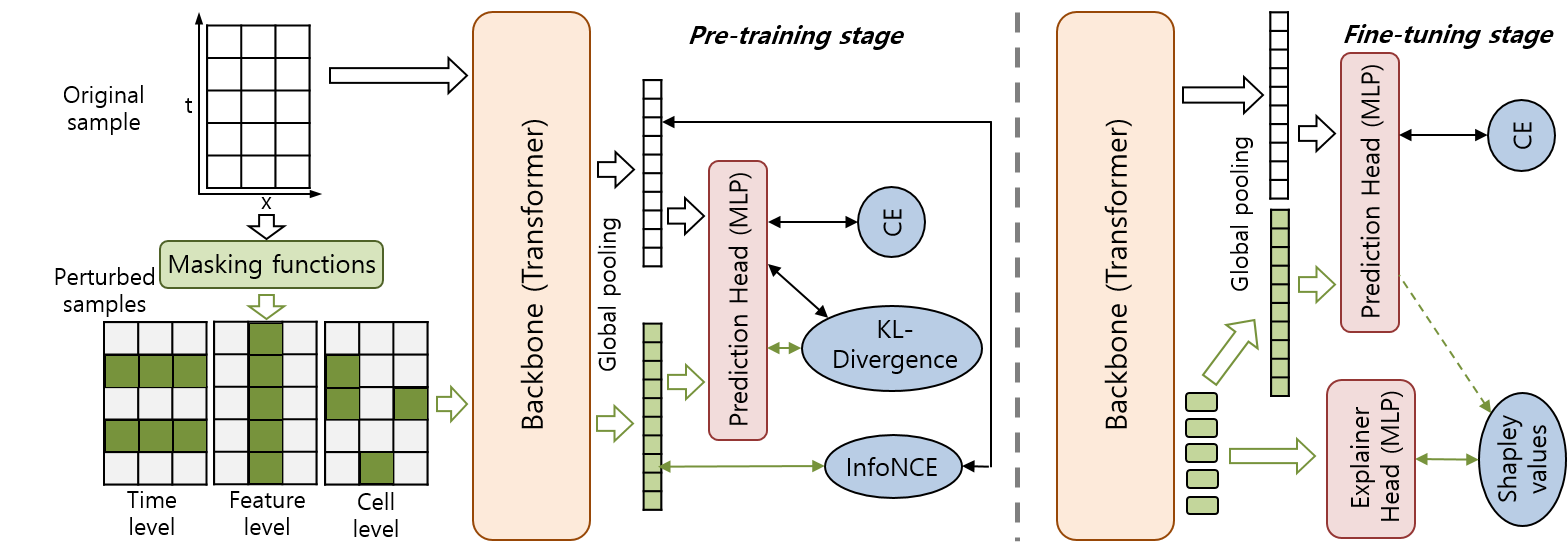}
\vspace{-0.1in}
\caption{The ShapTST framework. It involves 1) a Shaley-based pre-training stage, where TST is trained on both original and perturbed data to learn robust representations and capability to predict with partial information, and 2) a fine-tuning stage that enables TST to generate predictions alongside explanations.}
\label{fullarch}
\vspace{-0.1in}
\end{figure*}

In this paper, we present ShapTST (Figure~\ref{fullarch}), a novel framework that addresses the aforementioned needs in time-series modeling by unifying prediction and Shapley-value-based explanation generation within a single time-series transformer model. Unlike conventional approaches that depend on computationally intensive post-hoc explanation methods, ShapTST incorporates Shapley-based explanation mechanisms directly into the model during training. Through a novel Shapley-based pre-training strategy inspired by the FastSHAP framework \cite{jethani2021fastshap}, we eliminate expensive inference-time computations by replacing them with a one-time pre-training step conducted during the training phase. Essentially, this Shapley-based pre-training emulates the repeated inference process, where different elements of a sample are removed, and trains the model to make predictions using partial information while estimating the contributions of the missing elements. To enable explanations at both feature and time-step granularities, we introduce a multi-level masking approach. Moreover, our Shapley-based pre-training shares similarities with contrastive learning formulations, which have been widely recognized for their ability to enhance downstream prediction performance \cite{tian2020makes}. This specialized design not only enables efficient explanations generation but also improves the prediction accuracy by encouraging the model to focus on critical features and events essential for accurate predictions. Furthermore, the incorporated Shapley-value mechanism can be leveraged to regularize the model to prioritize critical information, which is particularly valuable for handling noise that is prevalent in real-world time-series data.

Our extensive experiments demonstrate the effectiveness of the proposed ShapTST framework by showing competitive prediction results, significant improvement of efficiency in generating accurate Shapley values, and robustness against data noise on various time-series tasks, both in classification and regression. Our contributions are threefold: (1) To our best knowledge, we are the first to incorporate Shapley value estimation into the training of time-series transformer models, (2) our proposed ShapTST yields accurate prediction alongside explanations without massive post-hoc computational overhead, and (3) we demonstrate the improvement in model robustness to data noises with a controllable Shapley-based regularization. 

This paper is organized as follows: In Section 2, we present our proposed methodology in detail. In Section 3, we show the experimental results with a thorough discussion. In Section 4, we provide related work. We conclude our paper in Section 5.

\section{Methodology}
In this section, we present ShapTST, a model that aims to simultaneously achieve high prediction performance and generate efficient model explanations. To achieve this goal, we use TST as the backbone model, known for its high performance and efficient processing of time sequences \cite{vaswani2017attention}. The model architecture is shown in Figure \ref{fullarch}

We define the input data as ${x,y}\in X \times Y$, where $x\in R^{D\times T}$ represents the input sample, $y\in \{1,...K\}$ is the associated label, $D$ is the feature dimension, $T$ is the time, and $K$ is the number of classes. 

\subsection{ShapTST}
Compare to the TST, ShapTST has an extra explainer output head alongside the prediction head such that it can provide both explanation and prediction in one forward pass. The prediction head $f_w(x)\rightarrow y$ is designed to learn the true labels $y$ given $x$, with the loss function: $\mathcal{L}_{w}=CE(\widehat{y},y)$, where $CE(\widehat{y},y)$ is the cross-entropy loss between the predicted label $\widehat{y}$ and the true label $y$.

The explainer head is designed to learn a function $\phi_{\gamma}(x,y)\in R^{D\times T}$ that generates Shapley values for each cell in $x$ of class $y$, where $\gamma$ represents the explainer's header parameters. To accomplish this, we use a loss function similar to that of FastSHAP \cite{jethani2021fastshap}, which supervises the estimation according to the following loss:
\begin{equation}
\begin{split}
\mathcal{L}_{\gamma} = E_{p(x)} E_{U(y)} E_{p(S)} \big[
(v_{x,y}(S) - v_{x,y}(\emptyset) \\
- S^T\phi_{\gamma}(x,y))^2 \big]
\end{split}
\end{equation}

Here, $S$ represents the subset of support with perturbations, $p(S)$ is a distribution over all possible $S$ over time steps, features, and cells, and $U(y)$ is uniformly sampled from a distribution of $K$ classes. We co-optimize the TST backbone blocks, the prediction heads, and the explainer head on the original data: $ f_{w'}(f_{w\setminus w'}(x),\phi_{\gamma}(x))\rightarrow y$, where $\phi_{\gamma}(x) \in R^{T,D,K}$. The total loss function that we optimize towards is thus given by:

\begin{equation}
\mathcal{L}_{DT} = \sum_{i\in\left \{ T,D,C \right \}}\mathcal{L}_{\gamma}^{i} + \mathcal{L}_{w}
\end{equation}

Here, $T, D,$ and $C$ represent the sampling of the support along time steps, features, and cells.

During inference, our model simultaneously generates predictions and Shapley value estimations in a single forward pass. The final Shapley values are refined based on the output through the efficiency gap approach \cite{jethani2021fastshap}. We compute the Shapley values for each cell using the learned Shapley estimation function $\phi_{\gamma}(x,y)$ and then adjust them by adding the average difference between the value of the entire set of cells and the value of the empty set. Mathematically, this can be expressed as follows:
\begin{equation}
\begin{split}
\phi_{\gamma}^{'}(x,y) = & \phi_{\gamma}(x,y) + \frac{1}{D \times T} \big( 
v_{x,y}(\mathbbm{1}) - v_{x,y}(\emptyset) \\
& - \mathbbm{1}^T \phi_{\gamma}(x,y) \big)
\end{split}
\label{shapeq}
\end{equation}

Here, $\mathbbm{1}$ is the complete set of support (all features), $v_{x,y}(S)\rightarrow R$ represents the characteristic function of subset support $S$. It is worth noting that our approach offers a significant improvement in speed compared to previous methods \cite{bento2021timeshap} that estimate Shapley values separately for each sample by re-evaluating sampled cell subsets post-hoc.

\subsection{Mask Design}
\begin{figure}[h]
\vspace{-0.1in}
\centerline{\includegraphics[width=2.5in]{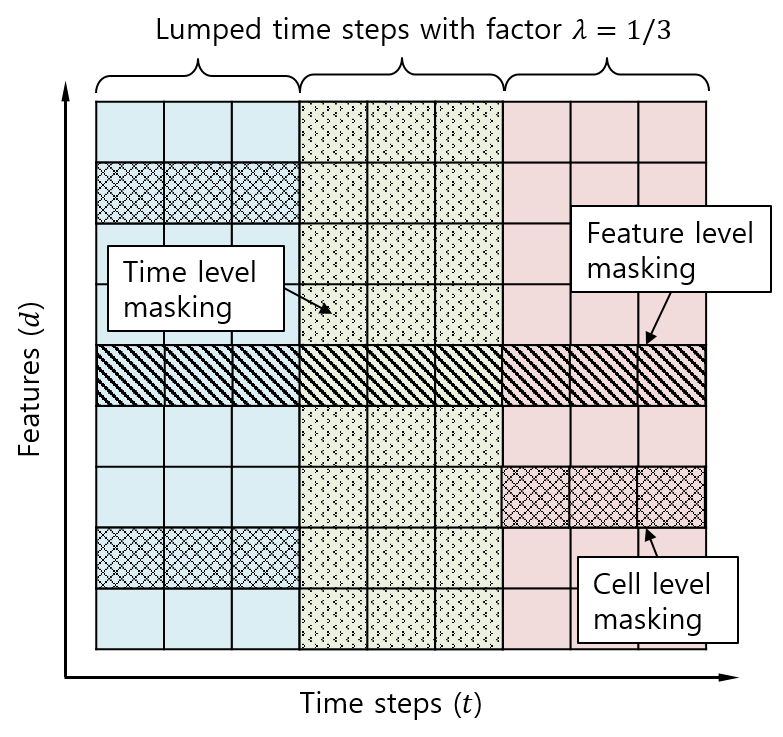}}
\vspace{-0.1in}
\caption{Maskings for the training of ShapTST}
\label{mask}
\end{figure}

Our approach employs perturbations through masking, where subsets of features or time steps are set to zero. Building on \cite{bento2021timeshap, park2019specaugment}, we design the masking function for time-series data along the time, feature, and cell levels, enabling explanations for events (time), features, and both. Masking creates local data dependencies, allowing the model to observe output changes when input subsets are masked. The explainer head uses this to estimate the importance of features or time steps for the model’s output. By using diverse masking functions, we generate explanations at varying granularities, aiding in understanding the model’s behavior across contexts.

\subsubsection{Time level masking}
A multivariate time series dataset can be represented by a matrix $x \in \mathbb{R}^{T \times D}$, and an accompanying matrix $\widetilde{x} \in \mathbb{R}^{T \times D}$ containing uninformative values. For the time dimension, a masking function perturbs the matrix $x$ by replacing a random subset of time steps $S_T \in \{0,1\}^{T}$ (i.e. columns in the matrix) with values from $\widetilde{x}$. In many real-world time series datasets, adjacent time steps contain similar values or are highly correlated. To account for this, we aggregate adjacent time steps using an aggregation factor $\lambda$. The masking function for the time level ($m_{T}$) is defined as follows:
\begin{equation}
m_{T}(x, S_{T},\lambda ) = tile_{\lambda}(S_T \cdot (x\cdot \lambda ) + (\mathbbm{1}-S_T)\cdot (\widetilde{x}\cdot \lambda ))
\end{equation}
Here, $tile_{\lambda}$ denotes the operation of tiling a matrix along the time axis with a factor of $\lambda$.

\subsubsection{Feature level masking}
The feature level masking function is used to replace a random subset of features with uninformative values in a given matrix $x$. Here, the subset of features is denoted as $S_D \in \{0, 1\}^D$, where $D$ represents the number of features. The matrix $\widetilde{x}$ has the same shape as $x$ and contains uninformative values. Since features are often independent, there is no need for aggregation along the feature dimension. The masking function $m_D$ can be expressed as:
\begin{equation}
m_{D}(x, S_{D}) = S_{D} \cdot x + (\mathbbm{1}-S_{D})\cdot \widetilde{x},
\end{equation}
where $\cdot$ denotes element-wise multiplication and $\mathbbm{1}$ is a vector of ones with the same shape as $S_D$.

\subsubsection{Cell level masking}
The matrix $x$ represents multivariate time series data, where each cell corresponds to a specific feature value at a particular time step. Although the model's explainability is mainly achieved through feature and time-level masking, cell-level perturbations are still included during training to ensure the model's robustness against various types of perturbations and to allow for cell-level inquiries during inference. Specifically, the cell-level masking function ($m_C$) is defined as:
\begin{equation}
m_{C}(x, S_{D},S_{T}) = (S_{T} \cdot S_{D}) \cdot x + (\mathbbm{1}-S_{T}\cdot S_{D})\cdot \widetilde{x}
\end{equation}
Here, $S_D$ and $S_T$ represent the subset of features and time steps to be masked, respectively. The resulting masked matrix $m_C(x, S_D, S_T)$ retains only the cell values where both the corresponding feature and time step are not masked.

\begin{table*}[htbp!]
\caption{Prediction Performance on Classification and Regression Tasks} \label{tab_clsperf}
\begin{center}
\begin{tabular}{@{}llccc@{}}
\toprule
\textbf{Task}                        & \textbf{Dataset}         & \textbf{TST} / \textbf{Shap-TST}       & \textbf{PatchTST / Shap-PatchTST} & \textbf{ViTST / Shap-ViTST}\\ 
\midrule
\multirow{4}{*}{\textbf{Binary (AUROC)}} 
                                     & AReM                     & 0.922 / \textbf{0.949}         & - / -         & \textbf{0.975} / 0.972 \\
                                     & EEG-eye                  & 0.757 / \textbf{0.779}         & - / -         & 0.822 / \textbf{0.825} \\
                                     & Gas sensor               & 0.939 / \textbf{0.966}         & - / -         & 0.960 / \textbf{0.964} \\
                                     & Maintenance              & 0.902 / \textbf{0.911}         & - / -         & \textbf{0.918} / 0.917 \\
\midrule
\multirow{2}{*}{\textbf{Multiclass (Macro-F1)}} 
                                     & AReM                     & 0.885 / \textbf{0.910}         & - / -         & \textbf{0.922} / 0.920 \\
                                     & Gas sensor               & 0.854 / \textbf{0.896}         & - / -         & 0.898 / \textbf{0.904} \\ 
\midrule
\multirow{2}{*}{\textbf{Regression (MSE)}} 
                                     & Benz. Concentration      & 2.053 / \textbf{2.057}                  & 1.978 / \textbf{1.977}         & - / - \\
                                     & L. Fuel Moisture         & 42.993 / \textbf{40.861}                 & 39.767 / \textbf{38.995}         & - / - \\
\bottomrule
\end{tabular}
\end{center}
\end{table*}

\subsection{Shapley-based pre-training}
We integrate the learning on partial information into the pre-training stage for TST ($f_w$), which emulates the Shapley value generation process. Specifically, we add perturbations to the data sample $x$ in time steps, features, and cells (i.e. intersections of time steps and features). Let ($\widetilde{\textbf{x}},\widehat{\textbf{x}}$) be two perturbed samples corresponding to $x$. To improve the model's representation learning, we leverage the generic contrastive pre-training approach as in SCARF \cite{bahri2021scarf}, using the InfoNCE loss. In addition to this, we supervise the prediction training using cross-entropy (CE) loss and minimize the KL-divergence between the two corresponding pseudo-soft labels ($\widehat{y},\widetilde{y}$).

To improve the Shapley value estimation, we incorporate the KL-divergence term, which encourages the model to make "best effort" predictions for perturbed samples, as suggested by \cite{jethani2021fastshap}. We combine the InfoNCE loss, the cross-entropy (CE) loss, and the KL-divergence loss into a weighted sum to form the overall pre-training loss. The weights $\alpha_{CE}$ and $\alpha_{KL}$ control the relative importance of the CE and KL-divergence losses compared to the InfoNCE loss. The pre-training loss of each batch is calculated as follows:
\begin{equation}
\begin{split}
\mathcal{L}_{P} = & \, InfoNCE(\widetilde{\textbf{x}}, \widehat{\textbf{x}}) 
+ \sum_{y \in B} \alpha_{CE} \cdot CE(\widehat{y}, y) \\
& + \alpha_{KL} \cdot KL(\widehat{y}, \widetilde{y})
\end{split}
\end{equation}
Here, $\widetilde{\textbf{x}}$ and $\widehat{\textbf{x}}$ are the representations of the original and perturbed versions of the input data, respectively. The InfoNCE loss improves the quality of learned representations, while the CE loss supervises the prediction head training. The KL-divergence loss minimizes the difference between the predictions of the original and perturbed versions of the input data, allowing for Shapley value estimation. The weights $\alpha_{CE}$ and $\alpha_{KL}$ control the relative contributions of the CE and KL-divergence losses to the pre-training loss.

\subsection{Shapley value-based regularization}
Real-world time-series prediction often suffers from adversarial noisy data \cite{karim2020adversarial, cheng202272}. To address this challenge and stabilize training by incorporating prior knowledge, we propose two regularizations for training TST.

Firstly, we introduce a mask probability during pre-training, which is determined by confidence measures such as variance or prediction error. This probability generates a binary mask matrix $M \in \{0,1\}^{T \times D}$, masking each entry with $p_{mask}$. The perturbed input $x' = M \cdot x + (1-M) \cdot \widetilde{x}$ is then used to enhance the model’s ability to handle noisy data. Secondly, we add a loss term in fine-tuning, minimizing the $L_2$ distance between estimated Shapley values $\phi$ and a controllable target $y_{target}$. This target, such as zero (indicating no effect), can be set based on prior knowledge or inferred from data subset, aligning Shapley value estimations with desired interpretations. Together, these regularizations enhance model’s robustness to noise and alignment with prior knowledge.




\section{Experiment}
We test our ShapTST approach on various time-series datasets: Activity Recognition Systems based on Multisensor data fusion (AReM) \cite{palumbo2016human}, EGG eye state detection \cite{wang2014eeg}, Gas Sensor Measurement for Indoor Activity Monitoring \cite{huerta2016online}, Intelligent Maintenance of Hydraulic System \cite{helwig2015condition}, and regression benchmark datasets from \cite{tan2020monash}. For a comprehensive evaluation, we perform the test on both binary classification and multi-class classification benchmarks, as well as some regression tasks. For binary classification, we transform the multi-class tasks to binary classification by selecting one class to predict. We use the Macro-F1 score for multi-class classification and the Area Under the Receiver Operating Characteristic curve (AUROC) as the evaluation metric for all binary classification analyses to alleviate the bias from label imbalance for a fair comparison.  For regression, we use the mean squared error (MSE) as the evaluation metric.

We benchmark our approach against three baseline transformers: 1) TST \cite{zerveas2021transformer}, a general-purpose Transformer for time-series classification and regression; 2) PatchTST \cite{nie2022time}, designed for long-range prediction; and 3) ViTST \cite{li2024time}, a latest SOTA model. We integrate our ShapTST framework into these models and compare their performance on classification and regression tasks. For TST, we use three transformer layers with six attention heads and a feature dimension of 128. A classification token is appended to input tokens and passed through a linear layer for predictions. Unlike TST, where input time-series are encoded via a linear projection, we use a 1-D conv-layer to embed each time step, aggregating multiple variables, with learnable positional encoding for both models. All training and testing are conducted on a single Nvidia A6000 GPU and Intel i9-7980 CPU, with 80\% of data for training and 20\% for testing.

\subsection{Model prediction performance}
As shown in table \ref{tab_clsperf}, our approach (i.e. Shap-*) achieves consistent performance improvements over the vanilla baselines for both classification and regression benchmarks. For binary classification, the improvement appears to be related to the size of the datasets. On the Maintenance dataset, which contains the least amount of samples and features among the 4 datasets, we achieve a similar performance as the baseline. On the larger Gas Sensor dataset, we obtain a relatively larger AUROC improvement of 2.7\%. The improvement is probably a result of the Shapley-based pre-training. For multi-class classification, it seems the macro-F1 improvements are even more significant. The performance improvement is mainly due to the extra effective pretraining. For the other two stronger baslines, PatchTST and ViTST, our approach achieved results on par with the original version, without further improvement. 

\subsection{Model prediction robustness}
\begin{figure}[h]
\centerline{\includegraphics[width=3in]{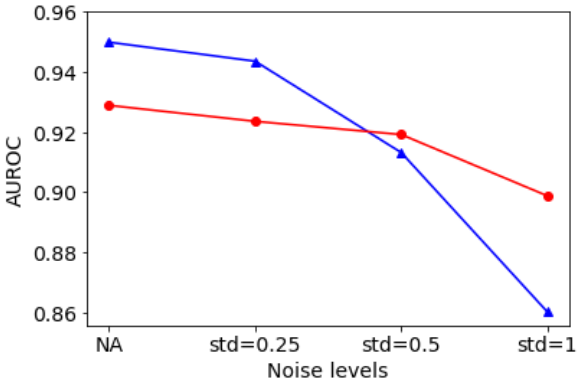}}
\vspace{-0.1in}
\caption{Performance on AReM dataset with added noise. The blue line is trained without Shapley-based regularization while the red line is trained with Shapley regularization} \label{noise_perf}
\vspace{-0.15in}
\end{figure}

\begin{figure}[h]
\vspace{-0.1in}
\centerline{\includegraphics[width=3in]{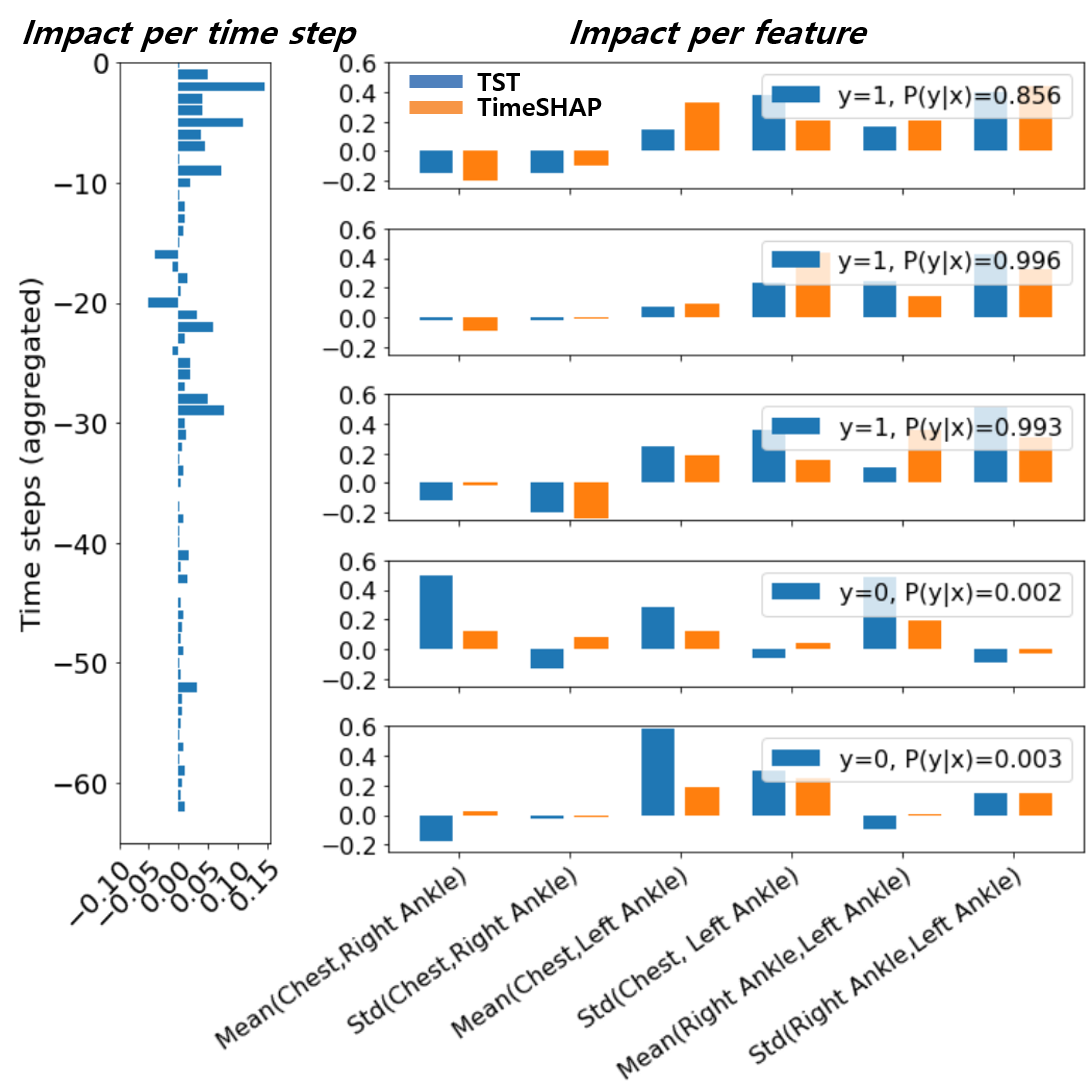}}
\caption{Time and feature level explanation examples (AReM). Adjacent 4 time points are aggregated into one time step.} \label{feature_arem}
\vspace{-0.1in}
\end{figure}

In order to showcase the robustness of our approach, we perform a further test on AReM dataset with added noise. Specifically, we added Gaussian noises with 0 mean and different standard deviation to one feature, std(Right Ankle, Left Ankle), which is important in the prediction of "cycling" activity (see Fig. \ref{feature_arem} ). To control the robustness against the noise, we set the  Shapley value of the feature to be 0 and penalize the model when it predicts a large corresponding Shapley value. In this way, we try to regularize the model learning by encouraging it to pay less attention to the noisy feature. Fig.\ref{noise_perf} shows that Shapley-based regularization makes model performance much more stable under strong noises, at the cost of modest performance sacrifice at weak noise.

\begin{table}[]
\begin{center}
\caption{Evaluation time to generate explanations} \label{tab_speedperf}
\begin{tabular}{@{}lll@{}}
\toprule
Dataset     & TST(TimeSHAP)   & ShapTST \\ \midrule
AReM        & 32.25 & 7.32    \\
EEG-eye     & 45.93 & 9.72    \\
Gas sensor  & 48.65 & 10.95   \\
Maintenance & 14.32 & 4.15    \\ \bottomrule
\end{tabular}
\end{center}
\end{table}

\begin{figure}[h]
\vspace{-0.1in}
\centerline{\includegraphics[width=3in]{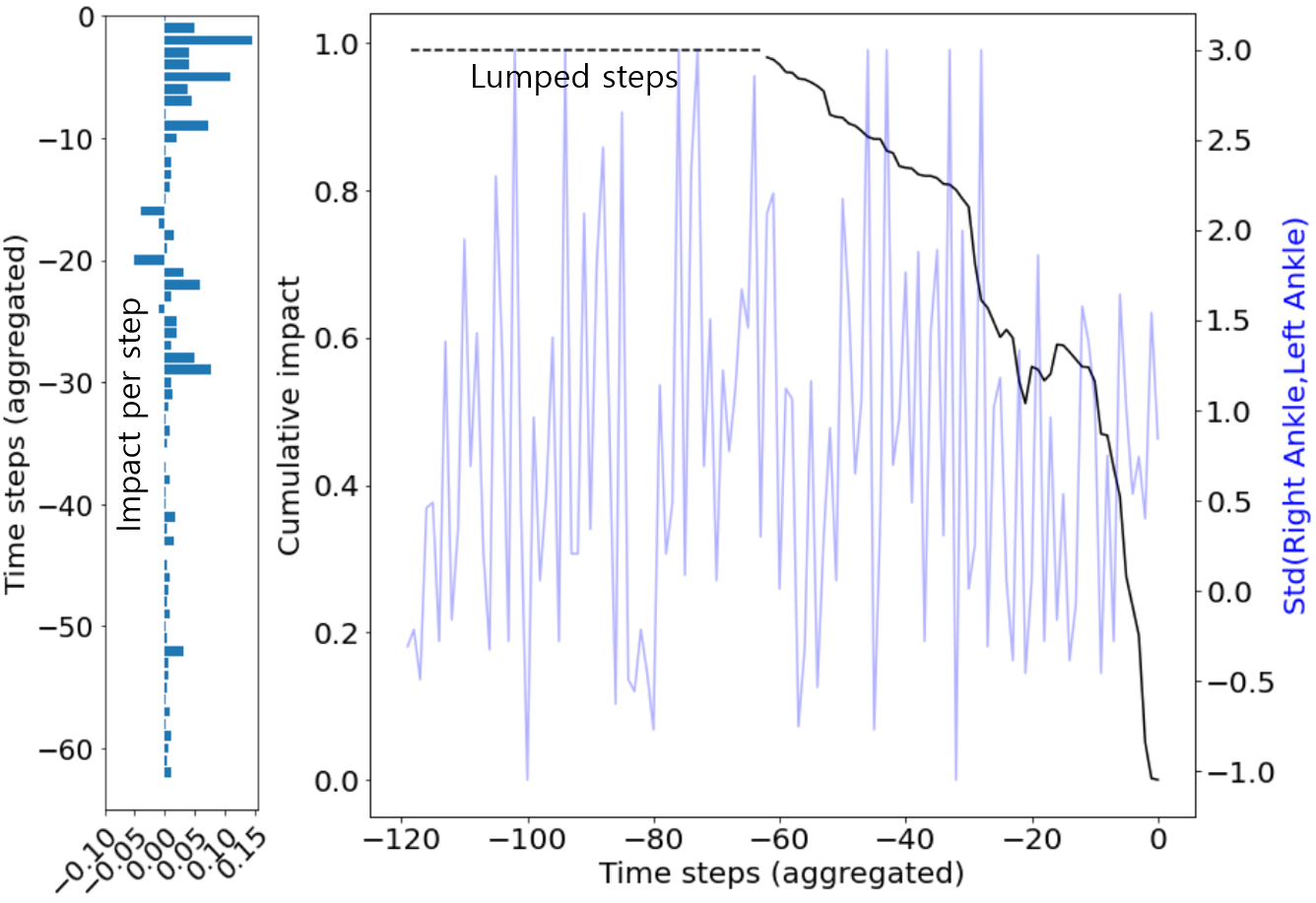}}
\vspace{-0.1in}
\caption{Time level explanation example of AReM Dataset. Adjacent 4 time points are aggregated into one time step.} \label{time_step_arem}
\end{figure}

\subsection{Model explanations}
\subsubsection{Time level explanation}
Figures \ref{time_step_arem} and \ref{feature_arem} illustrate time and feature-level quantitative explanations from the explainer head on the AReM dataset, showing the impact on model predictions in terms of magnitude and polarity. For time-level explanations, magnitude represents feature importance, while polarity indicates the direction in which a feature influences the model given specific values. Following \cite{bento2021timeshap}, we analyze the difference in impact between recent and older time steps by sequentially applying masks that cover increasing time steps from the start, generating predictions in a single forward pass, and calculating aggregated impacts for covered and uncovered steps using equation \eqref{shapeq}. Figure \ref{time_step_arem} highlights that recent time steps significantly influence activity predictions, with older steps being less relevant, consistent with \cite{bento2021timeshap}. This aligns with the data’s trend, where the most recent half, covering a typical period, dominates the impact (blue line in Fig. \ref{time_step_arem}).

\subsubsection{Feature level explanation}
Similarly, feature level explanations show the overall correspondence between features and its impact considering all time steps. For example in Fig.\ref{time_step_arem}, the movement of the left/right ankle strongly indicates that the person is cycling whereas the chest movement points to the opposite. On the positive samples, the estimated Shapley values are close to those of TimeSHAP.

\subsubsection{Explanation quality}
To evaluate the quality of generated explanations, we compare the faithfulness \cite{bhatt2020evaluating} of our generated explanations with those from TimeSHAP \cite{bento2021timeshap}. Table \ref{faith} reports the faithfulness scores averaged over the time steps and features for all 4 classification datasets, showing the quality of our explanations are on par with or better than TimeSHAP.

\begin{table}[]
\caption{Faithfulness of explanations} \label{faith}
\begin{center}
\begin{tabular}{@{}lll@{}}
\toprule
Datasets    & TimeSHAP & ShapTST \\ \midrule
AReM        & 0.650    & \textbf{0.688}   \\
EEG-eye     & \textbf{0.378}    & 0.369   \\
Gas sensor  & 0.889    & \textbf{0.890}   \\
Maintenance & 0.937    & 0.935   \\ \bottomrule
\end{tabular}
\end{center}
\end{table}

\subsection{Model explanation efficiency}
Table \ref{tab_speedperf} compares the speed for explanation generation between our method and TimeSHAP. It shows that our approach achieves a 4-5X reduction of evaluation time on all test datasets. This is probably a consequence of amortized estimation of Shapley values,  explanation generation in a single forward pass, and the transformer architecture.

\section{Conclusion}
We introduce ShapTST, a novel time-series transformer (TST) framework that integrates Shapley-based explanations. By combining amortized Shapley value estimation with pre-training, it enables valid explanations without extra computational cost post-inference, enhances classification and regression performance, and improves robustness against noise. Extensive experiments on public datasets demonstrate its competitive performance and explainability, advancing explainable deep learning for time-series analysis in safety-critical applications. With this work, we hope to pave the way for explainable and efficient deep time-series model for safety-critical applications.



\bibliographystyle{ieeetr}
\bibliography{conference_101719}

\end{document}